\documentclass[runningheads]{llncs}
\usepackage[T1]{fontenc}
\usepackage{graphicx}
\usepackage{footmisc}
\usepackage{url}
\usepackage{booktabs}
\usepackage{array}
\usepackage{amsmath}
\usepackage{orcidlink}
\renewcommand{\orcidID}[1]{\,\orcidlink{#1}}

\usepackage{pifont}
\newcommand{\checkmark}{\ding{51}}
\newcommand{\xmark}{\ding{55}}
\usepackage{bbding}
\usepackage{hyperref}
\usepackage{color}

\usepackage{float}

\begin{document}
\mainmatter

\title{Learning on the Fly: Replay-Based Continual Object Perception for Indoor Drones}
\titlerunning{Replay-Based Continual Object Perception for Indoor Drones} 

\author{
  Sebastian-Ion Nae\inst{1}\orcidID{0009-0008-6300-876X}\Envelope \and 
  Mihai-Eugen Barbu\inst{1}\orcidID{0009-0001-4063-0272} \and 
  Sebastian Mocanu\inst{1}\orcidID{0009-0007-0313-4724} \and 
  Marius~Leordeanu\inst{1,2,3}\orcidID{0000-0001-8479-8758}
}
\authorrunning{Sebastian-Ion Nae et al.}
\tocauthor{Sebastian-Ion Nae, Mihai-Eugen Barbu, Sebastian Mocanu, Marius Leordeanu}

\institute{
National University of Science and Technology Politehnica Bucharest, Romania
\and
Institute of Mathematics ``Simion Stoilow'' of the Romanian Academy, Romania
\and
NORCE Norwegian Research Center, Norway
\email{sebastian\_ion.nae@stud.fils.upb.ro, mihai\_eugen.barbu@stud.acs.upb.ro, sebastian.mocanu@upb.ro, leordeanu@gmail.com}
}

\maketitle

\begin{abstract}
Autonomous agents such as indoor drones must learn new object classes in real-time while limiting catastrophic forgetting, motivating Class-Incremental Learning (CIL). However, most unmanned aerial vehicle (UAV) datasets focus on outdoor scenes and offer limited temporally coherent indoor videos. We introduce an indoor dataset of $14,400$ frames capturing inter-drone and ground vehicle footage, annotated via a semi-automatic workflow with a $98.6\%$ first-pass labeling agreement before final manual verification. Using this dataset, we benchmark 3 replay-based CIL strategies: Experience Replay (ER), Maximally Interfered Retrieval (MIR), and Forgetting-Aware Replay (FAR), using YOLOv11-nano as a resource-efficient detector for deployment-constrained UAV platforms.
Under tight memory budgets ($5-10\%$ replay), FAR performs better than the rest, achieving an average accuracy (ACC, $mAP_{50-95}$ across increments) of $82.96\%$ with $5\%$ replay. Gradient-weighted class activation mapping (Grad-CAM) analysis shows attention shifts across classes in mixed scenes, which is associated with reduced localization quality for drones. The experiments further demonstrate that replay-based continual learning can be effectively applied to edge aerial systems. Overall, this work contributes an indoor UAV video dataset with preserved temporal coherence and an evaluation of replay-based CIL under limited replay budgets.  Project page: \href{https://spacetime-vision-robotics-laboratory.github.io/learning-on-the-fly-cl}{https://spacetime-vision-robotics-laboratory.github.io/learning-on-the-fly-cl}

\end{abstract}

\section{Introduction}
Object detection underpins autonomous systems in industrial inspection \cite{liu2024uav}, warehousing \cite{jordan2018state}, tracking \cite{jiang2021anti}, real-time visual servoing \cite{Mocanu_2025_ICCV}, and marine surveying \cite{wen2018unmanned}. Modern real-time detectors (e.g., YOLO variants and transformer-based RT-DETR \cite{wang2024rtdetrv3realtimeendtoendobject}) achieve speed-accuracy trade-offs, but they are trained and evaluated under closed-set assumptions. Benchmarks such as ImageNet-1k \cite{kisel2024flawsimagenetcomputervisions} and COCO~\cite{lin2015microsoftcococommonobjects} fix the label space, whereas robots should handle evolving class inventories and distribution shift \cite{neuwirth2025rico}. This motivates class-incremental learning for detection: models are trained over a sequence of tasks that introduce new categories while mitigating catastrophic forgetting of previous classes~\cite{wang2021wanderlust,kirkpatrick2017overcoming}.

Indoor unmanned aerial vehicles (UAVs) exemplify this setting; they must incorporate new mission-specific targets (e.g., tools, signage, other drones) using compact on-board or edge models with tight compute and memory budgets. However, UAV benchmarks capture outdoor scenes and emphasize ground-plane categories (e.g., pedestrians and vehicles), with limited focus on temporally coherent indoor sequences or inter-drone interactions (e.g, VisDrone~\cite{zhu2018visionmeetsdroneschallenge} and UAVDT-style \cite{du2018unmannedaerialvehiclebenchmark} benchmarks). Related benchmarks study aerial single-object tracking or anti-UAV detection in the wild \cite{mueller2016benchmark,jiang2021anti}, but they are not designed for continual indoor object detection. Indoor UAV datasets do exist, targeting navigation and localization rather than continual object detection, as shown in Table \ref{tab:comp_datasets}, and lack class-evolving protocols required to study CIL~\cite{drones6080202,shalaby2025miluvmultiuavindoorlocalization}.

We introduce a drone-centric continual learning framework built on a temporally coherent indoor video dataset ($14,400$ frames) containing drone-to-drone and drone-to-ground-vehicle interactions. To reduce annotation cost, a frozen vision-language teacher (GroundingSAM \cite{ren2024groundedsamassemblingopenworld}) produces box/mask pseudo-labels that supervise a lightweight student detector (YOLOv11-nano \cite{jegham2024yolo}) suitable for deployment. The auto-annotation pipeline attains $98.6\%$ first-pass pseudo-label agreement prior to manual verification. We then evaluate three replay-based CIL strategies under strict buffer budgets: Experience Replay (ER) \cite{rolnick2019experience}, a variant of Maximally Interfered Retrieval (MIR) \cite{aljundi2019online}, and Forgetting-Aware Replay (FAR) using a tight buffer. Finally, we connect offline metrics to behavior by deploying the learned model in a sequential target-tracking task, and we probe model attention with Grad-CAM \cite{selvaraju2020grad}, which shows human-dominant activations in mixed scenes that coincide with reduced drone localization quality. 

\section{Related Work}
Continual learning for object detection (CLOD) has been studied mostly with two-stage detectors such as Faster R-CNN~\cite{peng2020faster}, where methods modify components, add task-specific losses, or replay prior-task images using augmentations (e.g., Mix/Mosaic)~\cite{cermelli2022modeling,menezes2023continual,liu2023augmented}. In contrast, many real deployments favor one-stage detectors (e.g., YOLO-style models) for real-time inference. A recurring challenge in one-stage incremental detection is the instability and noise of regression outputs~\cite{mo2024bridge}, motivating approaches based on distillation or feature-preservation objectives. For example, distillation has been explored for fully convolutional one-stage detectors such as FCOS~\cite{tian2019fcosfullyconvolutionalonestage}.

Knowledge distillation (KD) methods like Elastic Response Distillation~\cite {feng2022overcomingcatastrophicforgettingincremental} mitigate forgetting by constraining detection heads (e.g., L2 distillation) and tailoring losses across phases. Complementary lines of work avoid direct output distillation and instead combine intermediate feature distillation with balanced replay to improve stability~\cite{de2024replay}. Replay-based strategies typically require storing a subset of past samples, but they are simple to implement and are often reported to be effective at limiting catastrophic forgetting in practice.

Most CLOD evaluations rely on static, diverse image datasets, providing limited coverage of drone viewpoints. Drone-centric datasets such as Dogfight~\cite{ashraf2021dogfight} and RT Flying OD~\cite{reis2023realtime} partially address aerial perspectives, but they are outdoor-focused and follow standard batch-training protocols; moreover, prior work reports label quality limitations (e.g., missing/corrupted annotations) in these collections. RT Flying OD~\cite{reis2023realtime} further suggests that compact YOLO variants (e.g., YOLOv8-nano) offer favorable speed--accuracy trade-offs for edge deployment.

Motivated by constrained on-board learning and small replay buffers~\cite{aggarwal2023chameleon}, we introduce an indoor drone video dataset and use a frozen vision-language teacher to create pseudo-labels for a lightweight student detector (YOLOv11-nano). We then define a class-incremental detection setup where each class maps to a task and evaluate 3 replay-based strategies: Experience Replay (ER)~\cite{rolnick2019experience}, Maximally Interfered Retrieval (MIR)~\cite{aljundi2019online}, and a forgetting-aware replay heuristic (FAR, adapted from~\cite{aljundi2019online}), under strict memory budgets to assess suitability for UAVs.

\section{Approach}
We train YOLOv11-nano for indoor detection of drones and ground vehicles using (i) an indoor video dataset with semi-automatic annotations and (ii) a class-incremental learning pipeline with replay-based continual learning.

\subsection{Dataset}

To reduce the outdoor bias reported in ~\cite{reis2023realtime,ashraf2021dogfight}, we collected an \emph{indoor} aerial dataset in a controlled laboratory setting. Two human-piloted drones executed repeated circular trajectories around typical classroom objects, yielding four video streams: two on-board drone views and two third-person recordings (smartphone). We sampled each stream at 1 frame per second (FPS), producing $14,400$ frames at $3840 \times 2160$ resolution.

\begin{figure}[ht]
  \centering
    \includegraphics[width=\textwidth]{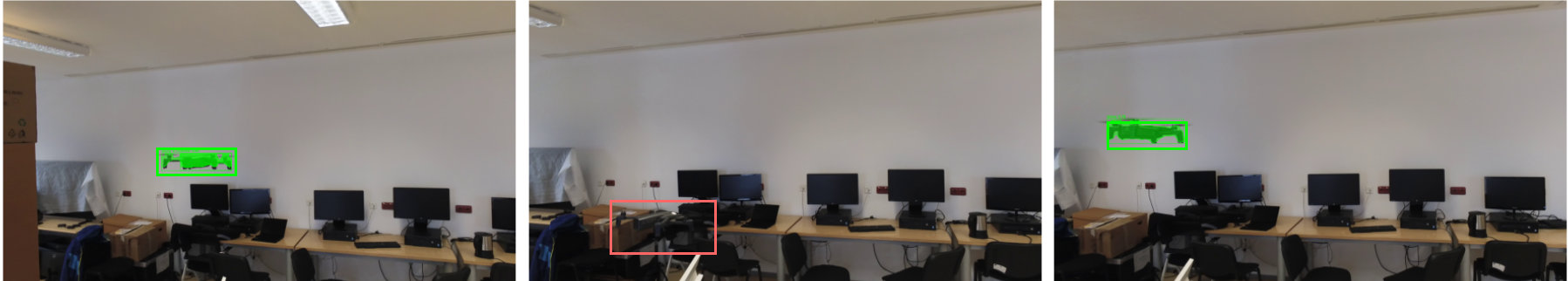}
    \caption{Example pseudo-label outcomes. Left/right: successful detections. Middle: a rare miss where GroundingSAM fails to localize a drone. Such cases occurred in approximately $1.4\%$ of frames and were corrected during human review.}
    \label{fig:miss_detection}
\end{figure}

To reduce manual labeling effort, we adopted a semi-automated labeling workflow. We run GroundingSAM~\cite{ren2024groundedsamassemblingopenworld} to predict instance masks and convert masks to axis-aligned bounding boxes. Predictions are filtered using a confidence threshold of $0.75$ and mask-to-box matching based on intersection-over-union (IoU) $\ge 0.5$. Prompts are kept minimal (e.g., \emph{``drone'': ``flying vehicle''}). We then conduct a full human review of the dataset: approximately $200$ out of $14,400$ frames required edits, shown in Figure \ref{fig:miss_detection}, i.e., \textbf{$98.6\%$} of frames were accepted without modification in the first pass.

\begin{table}[ht]
    \centering
    \caption{Dataset comparison with prior drone-centric detection datasets. We report whether the dataset provides indoor scenes and temporally coherent video sequences.}
    \begin{tabular}{l l c c c c c}
        \toprule
        \textbf{Paper} & \textbf{Dataset} & \textbf{\#Videos} & \textbf{\#Images} & \textbf{Indoor} & \textbf{Temporal} & \textbf{CIL} \\
        \midrule
        RT Flying OD~\cite{reis2023realtime} & Dataset 1 & 0 & 15,064 & \xmark & \xmark & \xmark \\
        RT Flying OD~\cite{reis2023realtime} & Dataset 2 & 0 & 11,998 & \xmark & \xmark & \xmark \\
        Dogfight~\cite{ashraf2021dogfight} & NPS-Drones & 14 & 70,250 & \xmark & \checkmark & \xmark \\
        Dogfight~\cite{ashraf2021dogfight} & FL-Drones & 50 & 38,948 & \xmark & \checkmark & \xmark \\
        \midrule
        \textbf{Ours} & \textbf{UAV-IndoorCL} & \textbf{4} & \textbf{14,400} & \checkmark & \checkmark & \checkmark \\
        \bottomrule
    \end{tabular}
    \label{tab:comp_datasets}
\end{table}

In addition to the drone video frames, we curate a ground-vehicle subset of $2,000$ images spanning three visually similar vehicle classes (fine-grained categories). This design (i) stress-tests replay under subtle inter-class differences (e.g., body profile and headlight geometry), (ii) supports realistic incremental updates where vehicle types are introduced sequentially, and (iii) induces a natural distribution shift across tasks. We define five tasks ordered as: ground vehicles (Tasks 1--3), aerial drones (Task 4), and humans (Task 5).

\subsection{Continual Learning for Object Detection}

Training proceeds over a task stream $\{\mathcal{T}_1,\dots,\mathcal{T}_5\}$ where each task introduces one new class. For task $\mathcal{T}_j$, the learner trains on the current task data plus a replay subset sampled from tasks $\{\mathcal{T}_1,\dots,\mathcal{T}_{j-1}\}$. After completing task $j$, we evaluate on the test split of every task $i \le j$ to measure (i) plasticity on the current class and (ii) retention of previous classes. We compare 5 training methods:

\begin{itemize}
    \item \textbf{Na\"ive fine-tuning.} Sequential training on each task using only current-task data (no replay/regularization), serving as a forgetting reference.
    \item \textbf{Experience Replay (ER)~\cite{rolnick2019experience}.} Uniformly samples a fixed-budget subset of prior-task images and merges them with the current task training set.
    \item \textbf{Maximally Interfered Retrieval (MIR)~\cite{aljundi2019online}.} Since the original MIR requires per-minibatch interference estimates, we use a lightweight proxy: from a capped candidate pool (default 800 prior images), we select the $K$ images (default $K{=}200$) with the lowest image-level detection recall@0.5 under the current model, and replay them for the next learning task.
    \item \textbf{Forgetting-Aware Replay (FAR).} After finishing task $s$, we cache image-level recall@0.5 for a subset of prior images using the task-$s$ checkpoint as baseline. Before learning task $t>s$, we rescore the same pool with the current model and prioritize replay by the recall drop.
    $\max(0,\text{recall}_{\text{baseline}} - \text{recall}_{\text{current}})$,
    selecting the top-$K$ images with the largest degradation.
    \item \textbf{Joint training.} Trains on all task training data simultaneously (uses future data), providing a performance ceiling, not a continual-learning method.
\end{itemize}

All methods share the same training configuration: AdamW optimizer; input size $640 \times 640$; HSV and small geometric augmentations; mosaic, mixup, copy-paste; and early stopping with patience 3 epochs. Image-level recall@0.5 uses greedy matching between predicted and ground-truth boxes at IoU $\ge 0.5$. Unless otherwise stated, YOLO inference uses a confidence threshold of $0.25$ and a non-maximum suppression (NMS) IoU threshold of $0.7$~\cite{khanam2024yolov11}. Replay pool caps and budgets are identical across ER/MIR/FAR for a controlled comparison.

\section{Experiments}

We evaluate replay-based continual learning methods (ER, MIR, FAR) under 4 replay buffer budgets: 5\%, 10\%, 25\%, and 50\% (following the protocol in~\cite{neuwirth2025rico}). We report COCO-style mean average precision $\text{mAP}_{50\text{-}95}$ for both bounding-box detection and instance segmentation. Continual learning metrics are computed on $\text{mAP}_{50\text{-}95}$: \emph{Average Accuracy} (ACC), the final average performance across tasks, and \emph{Backward Transfer} (BWT), which measures the impact of learning later tasks on earlier ones (negative values indicate forgetting).

\begin{table}[ht]
\centering
\caption{Continual learning results (ACC and BWT) across replay buffer budgets. Values are mean $\pm$ standard deviation over 3 seeds. Best per row are bolded.}
\label{tab:er_summary}
\scriptsize
\setlength{\tabcolsep}{2.5pt}
\resizebox{\textwidth}{!}{%
\begin{tabular}{l c c c c c c}
\toprule
& \multicolumn{3}{c}{\textbf{$\text{mAP}_{50\text{-}95}$ ACC $\uparrow$}} & \multicolumn{3}{c}{\textbf{$\text{mAP}_{50\text{-}95}$ BWT $\uparrow$}} \\
\cmidrule(lr){2-4}\cmidrule(lr){5-7}
\textbf{Buffer} & \textbf{ER} & \textbf{FAR} & \textbf{MIR} & \textbf{ER} & \textbf{FAR} & \textbf{MIR} \\
\midrule
\multicolumn{7}{l}{\textbf{Box Detection}} \\
5\%  & $61.18\pm8.18$ & $\mathbf{82.96\pm4.18}$ & $78.43\pm3.45$ & $-11.65\pm10.44$ & $\mathbf{-5.28\pm5.25}$ & $-9.47\pm2.09$ \\
10\% & $75.07\pm6.77$ & $\mathbf{86.48\pm3.00}$ & $82.24\pm3.73$ & $-3.44\pm4.99$ & $\mathbf{-1.42\pm1.83}$ & $-5.61\pm4.45$ \\
25\% & $85.11\pm2.52$ & $\mathbf{87.86\pm0.18}$ & $85.61\pm1.88$ & $2.84\pm3.61$ & $\mathbf{3.21\pm0.53}$ & $2.03\pm1.64$ \\
50\% & $\mathbf{88.69\pm1.02}$ & $87.92\pm0.45$ & $87.30\pm0.43$ & $\mathbf{3.66\pm0.41}$ & $3.64\pm1.05$ & $2.40\pm1.07$ \\
\addlinespace
\multicolumn{7}{l}{\textbf{Instance Segmentation}} \\
5\%  & $56.93\pm5.66$ & $\mathbf{77.74\pm3.72}$ & $74.39\pm4.60$ & $-11.76\pm6.98$ & $\mathbf{-4.22\pm4.58}$ & $-7.51\pm2.99$ \\
10\% & $69.18\pm2.76$ & $\mathbf{79.61\pm2.87}$ & $77.08\pm1.65$ & $-4.09\pm5.45$ & $\mathbf{-2.29\pm3.16}$ & $-5.08\pm2.90$ \\
25\% & $78.59\pm2.62$ & $\mathbf{80.37\pm0.09}$ & $78.43\pm0.82$ & $\mathbf{2.87\pm3.81}$ & $1.02\pm0.83$ & $0.19\pm1.57$ \\
50\% & $\mathbf{80.87\pm0.30}$ & $80.59\pm0.82$ & $79.71\pm0.84$ & $\mathbf{2.83\pm1.08}$ & $2.54\pm2.00$ & $2.18\pm1.29$ \\
\bottomrule
\end{tabular}
}
\end{table}

\begin{figure}[ht]
    \centering
    \begin{minipage}[b]{\textwidth}
        \centering
        \includegraphics[width=0.8\textwidth]{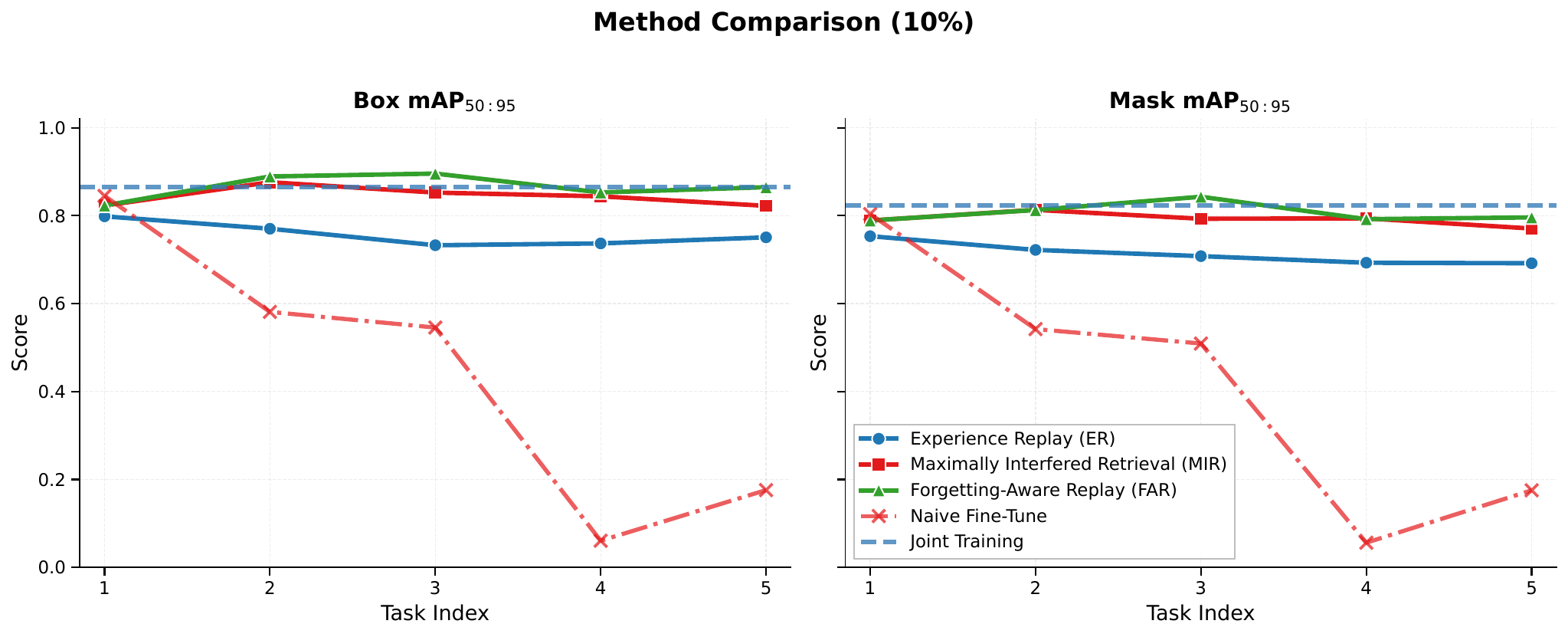}
        \label{fig:method_comparison_10p}
    \end{minipage}%
    \hfill
    \begin{minipage}[b]{\textwidth}
        \centering
        \includegraphics[width=0.8\textwidth]{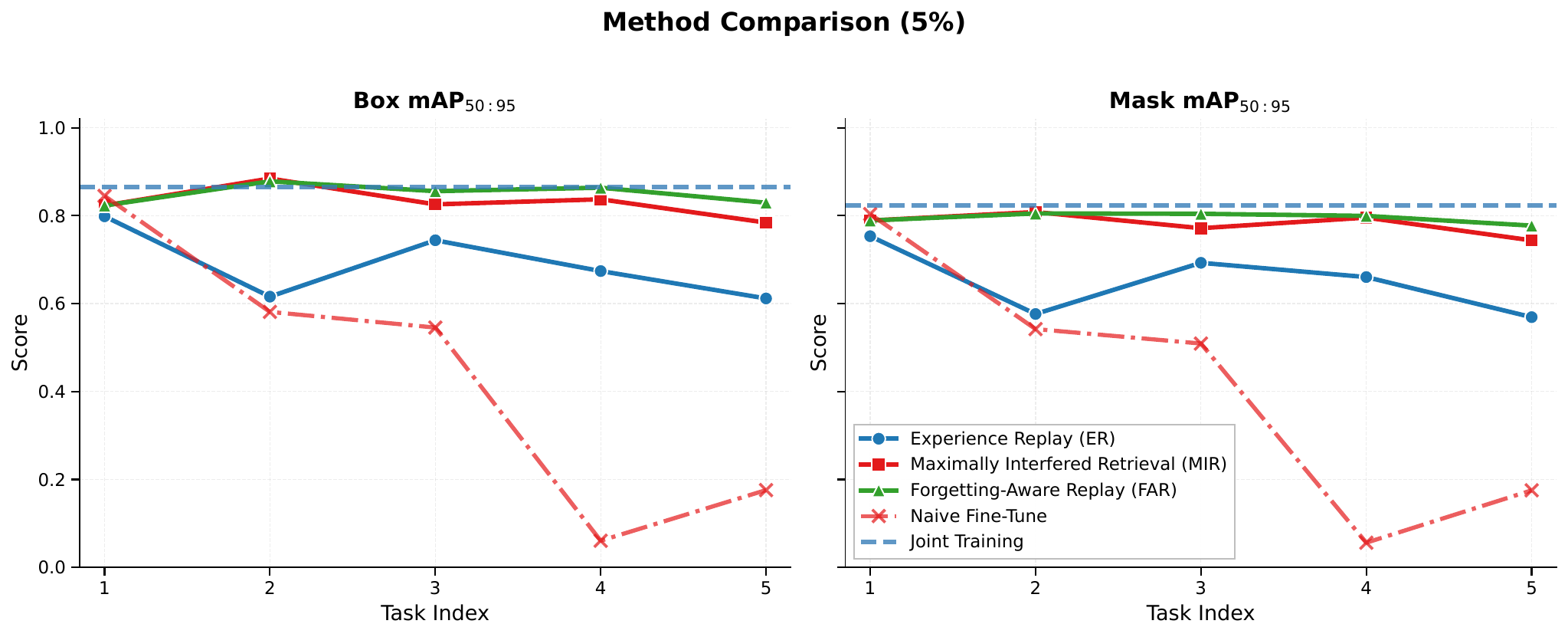}
        \label{fig:method_comparison_5p}
    \end{minipage}
    \caption{Task-wise $\text{mAP}_{50\text{-}95}$ for box detection and instance segmentation at $10\%$ (top) and $5\%$ (bottom) replay. Na\"ive fine-tuning shows forgetting on earlier tasks, while replay-based methods improve retention. FAR and MIR remain closest to the joint-training upper bound under these settings.}
    \label{fig:combined_comparison}
\end{figure}

Table~\ref{tab:er_summary} summarizes ACC and BWT across buffer sizes; the na\"ive baseline, which performs substantially worse, is shown only in Figure \ref{fig:combined_comparison}. FAR consistently leads at small budgets, while all replay methods converge at $25-50\%$.

Figure~\ref{fig:combined_comparison} illustrates performance trajectories for continual learning methods under low replay buffers of $5\%$ and $10\%$. The na\"ive baseline demonstrates clear catastrophic forgetting, while replay-based methods successfully preserve performance on earlier tasks. FAR and MIR emerge as better performers than the rest, maintaining accuracy near the joint training upper bound even with minimal memory allocation. 

Inference on an RTX 4060 Ti achieves $6.73$ ms per image, while CPU-only execution yields $51.27$ ms on a Ryzen 3700X ($4.4$ GHz) and $207.15$ ms on a Raspberry Pi 5 (Arm Cortex-A76, $2.4$ GHz, 4 GB RAM). Reported latencies are means over 30 runs, indicating the model can support high-frequency, real-time use.

\subsection{Grad-CAM Analysis}
\begin{figure}[ht]
  \centering
  \begin{minipage}[t]{0.48\textwidth}
    \includegraphics[width=\textwidth]{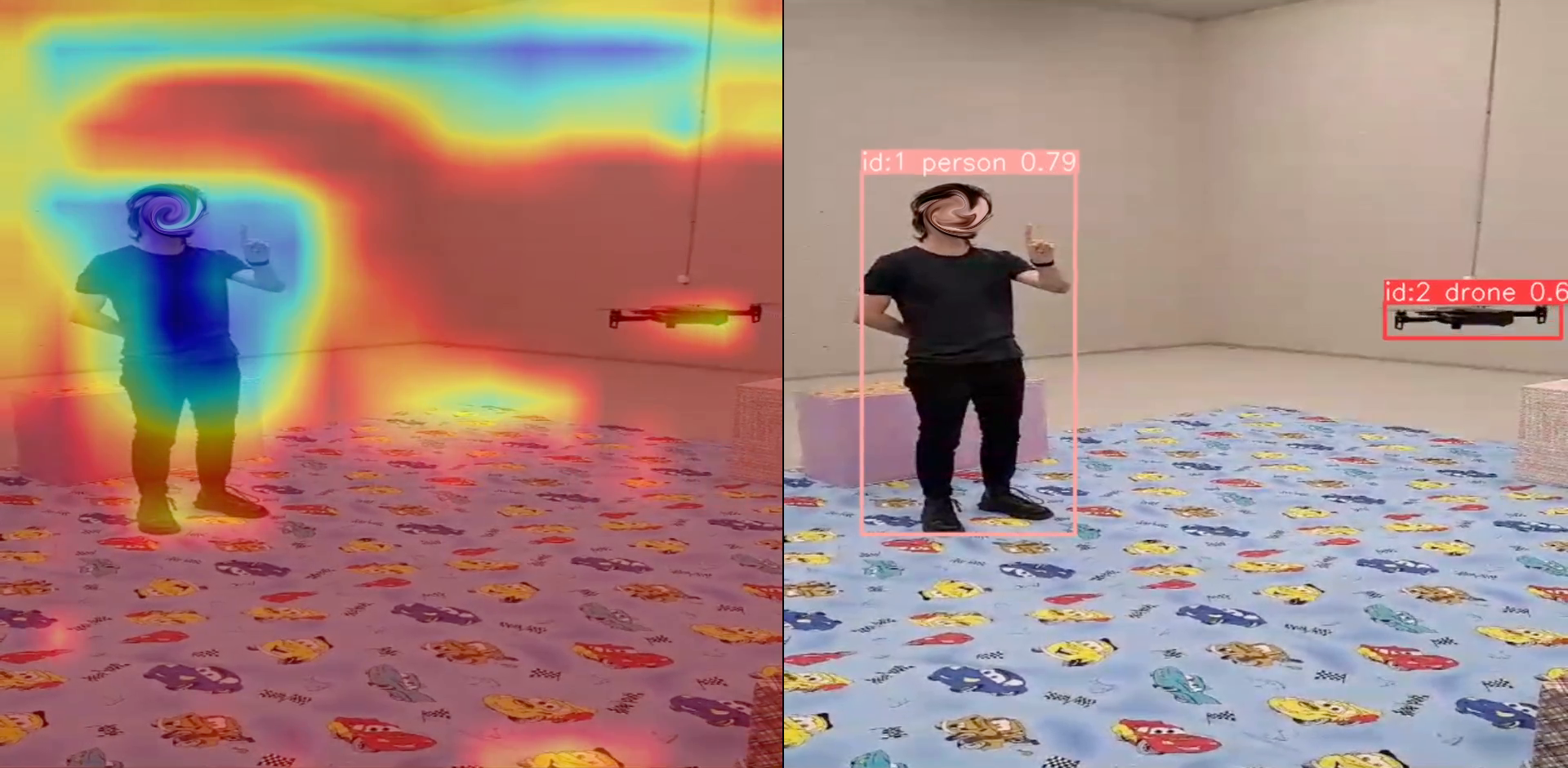}
    \caption{Grad-CAM highlighting a human attention pattern in mixed scenes.}
    \label{gradCAM}
  \end{minipage}
  \hfill
  \begin{minipage}[t]{0.48\textwidth}
    \includegraphics[width=\textwidth]{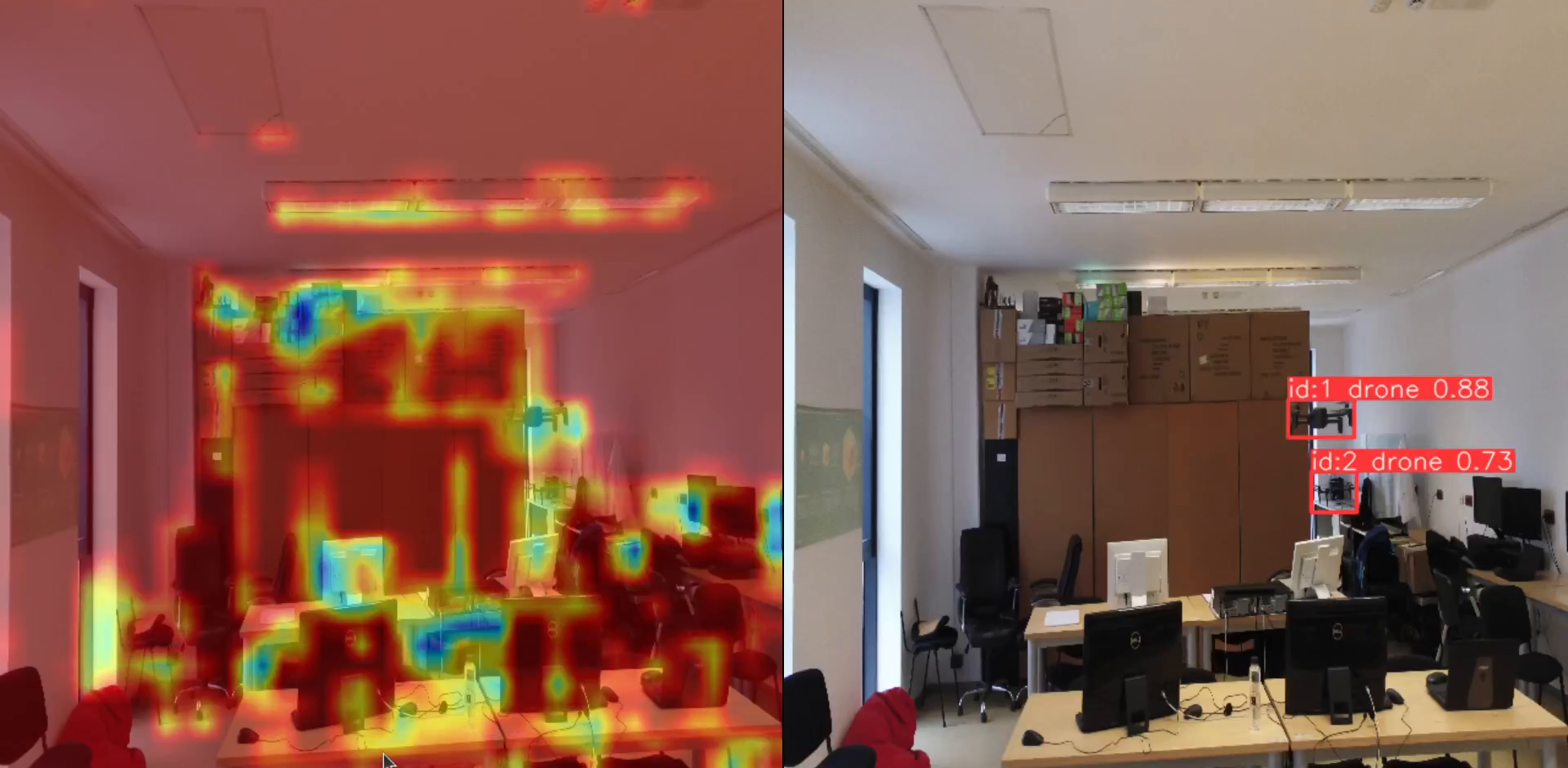}
    \caption{Grad-CAM showing distributed, target-focused attention in UAV-only scenes.}
    \label{gradCAM_drone}
  \end{minipage}
\end{figure}

We apply Grad-CAM~\cite{selvaraju2020grad} to visualize YOLOv11 attention patterns. In mixed human-drone scenarios (Figure~\ref{gradCAM}), attention concentrates on humans. Drone-only scenes (Figure~\ref{gradCAM_drone}) exhibit distributed attention across multiple targets with reduced background interference.

\begin{figure}[H]
  \centering
    \includegraphics[width=\textwidth]{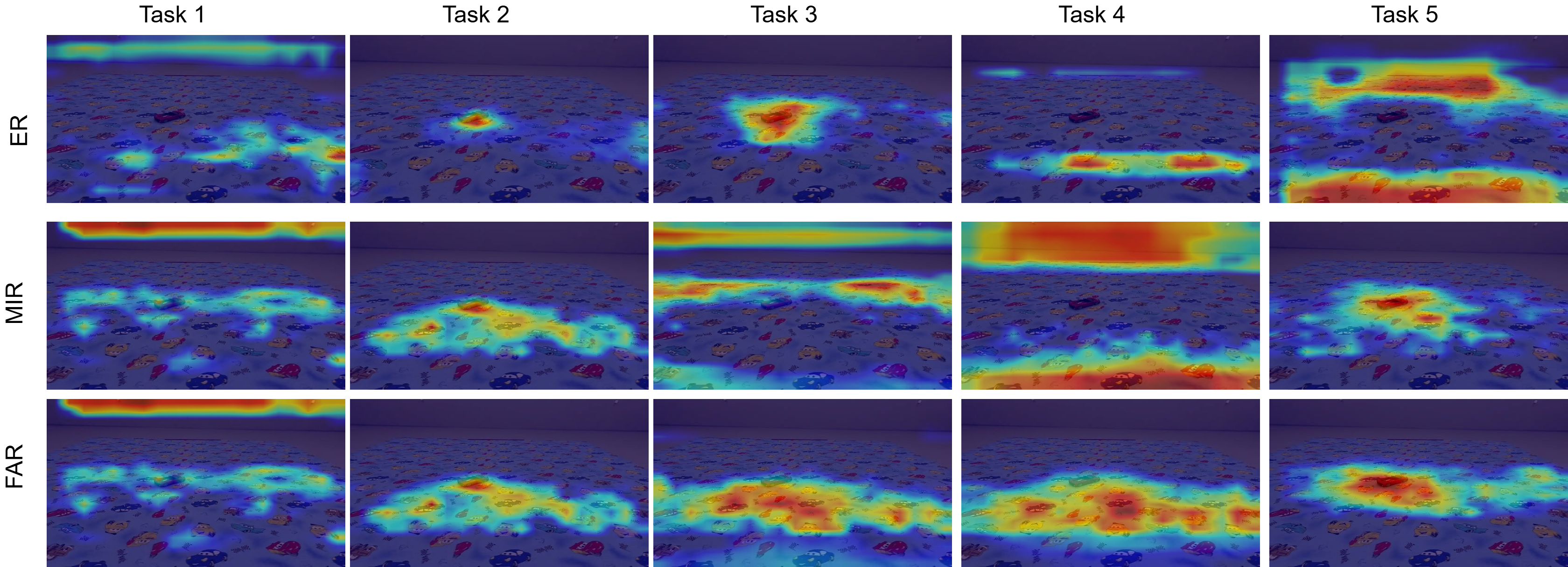}
    \caption{Grad-CAM visualizations of the final conv layer across five sequential tasks at 5\% replay buffer. Rows show ER, MIR, and FAR attention patterns for each task (warmer = higher activation). ER's attention degrades after Task 3 and becomes diffuse by Task 5, indicating poor retention. MIR and FAR maintain localized attention patterns across tasks, demonstrating plasticity and generalization. FAR is more concentrated, especially in Tasks 3-4, suggesting more discriminative features than MIR's broader attention.}
    \label{gradCAM-anal}
\end{figure}

These qualitative observations mirror the quantitative results: with tight memory, human-centric clutter increases interference, whereas drone-only scenes yield cleaner multi-target attention and more robust tracking. Figure~\ref{gradCAM-anal} shows distinct attention dynamics across replay methods: ER deteriorates after Task3, becoming nearly uniform by Task5, while MIR and FAR preserve localized, task-relevant saliency. FAR is the most concentrated (Tasks~3-4), consistent with its superior performance at 5\% replay.

\subsection{Qualitative Results}

\begin{figure}[ht]
  \centering
    \includegraphics[width=\textwidth]{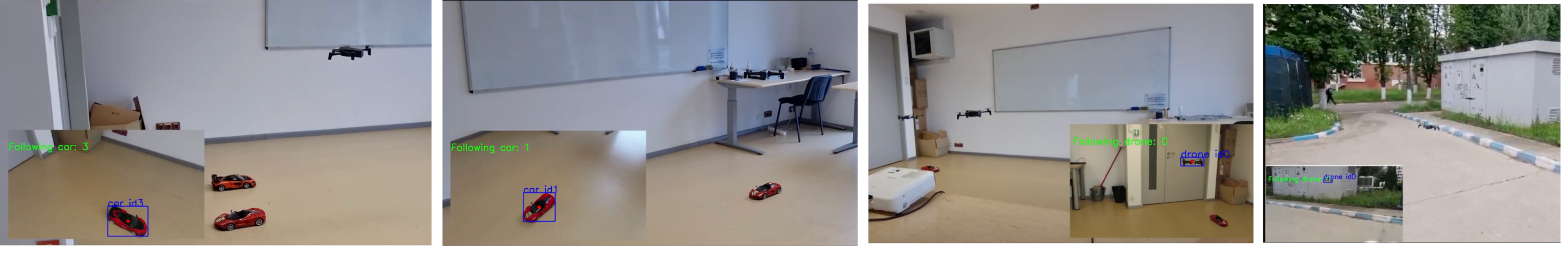}
    \caption{Qualitative examples of YOLOv11-based target following across diverse environments. The UAV tracks ground vehicles and aerial drones in indoor classrooms, constrained spaces, and outdoor scenes, demonstrating cross-environment robustness despite training exclusively on indoor data with textured flooring.}
    \label{fig:qualitative}
\end{figure}

We validate the FAR model (5\% replay) in environments not seen during training: (i) an indoor laboratory without the textured carpet present in the training data and (ii) outdoor settings for drone/person tracking. A video demonstration is available on the \href{https://spacetime-vision-robotics-laboratory.github.io/learning-on-the-fly-cl}{project page}. For closed-loop tracking, we use a proportional--integral--derivative (PID) controller to center the selected target in the camera frame. Figure~\ref{fig:qualitative} shows flight samples demonstrating dynamic target switching.

\section{Conclusion and Discussion}
We present a temporally consistent indoor UAV video dataset ($14,400$ frames) and a class-incremental benchmark for resource-constrained aerial perception. With replay buffers ranging from $5-50\%$, we compare three replay strategies (ER, MIR, FAR) and find FAR performs better at constrained budgets ($5-10\%$), reaching $82.96\%$ $\text{mAP}_{50\text{-}95}$ ACC with $5\%$ replay. At $25\%$ replay, results converge and BWT becomes non-negative, indicating moderate buffers recover most performance. We also demonstrate closed-loop tracking with the learned model, supporting practical onboard use. Grad-CAM visualizations highlight a failure in human-drone scenes, where attention often shifts to humans, coinciding with poorer drone localization, suggesting the need for mechanisms that curb attention drift in clutter.

\textbf{Limitations.} Our evaluation uses five classes and controlled indoor data; real deployments involve larger label spaces, more diverse viewpoints, and stronger domain shifts. The pseudo-labeling pipeline ($98.6\%$ first-pass agreement) still introduces residual noise that may accumulate across incremental steps. Finally, our tracking controller is a simplified PID baseline rather than a full visual-servoing stack. 

\textbf{Future work.} Promising directions include: (1) scaling to $20$+ categories under severe replay constraints, (2) adapting continual detectors across indoor-outdoor transitions, (3) hybrid replay--distillation that allocates memory to samples with high measured forgetting, and (4) attention-aware replay or regularization to mitigate human-dominant saliency in mixed scenes.

\begin{credits}
\subsubsection{\ackname}
Supported in part by projects ``Romanian Hub for Artificial Intelligence - HRIA'', Smart Growth, Digitization and Financial Instruments Program, 2021-2027 (MySMIS no. 334906), European Health and Digital Executive Agency (HADEA) through DIGITWIN4CIUE (Grant No. 101084054), and ``European Lighthouse of AI for Sustainability - ELIAS'', Horizon Europe program (Grant No. 101120237).

\subsubsection{\discintname}
The authors have no competing interests to declare that are relevant to the content of this article.
\end{credits}

\bibliographystyle{splncs04}
\bibliography{refs}

\begin{thebibliography}{10}
\providecommand{\url}[1]{\texttt{#1}}
\providecommand{\urlprefix}{URL }
\providecommand{\doi}[1]{https://doi.org/#1}

\bibitem{aggarwal2023chameleon}
Aggarwal, S., Binici, K., Mitra, T.: Chameleon: Dual memory replay for online continual learning on edge devices. IEEE Transactions on Computer-Aided Design of Integrated Circuits and Systems  \textbf{43}(6),  1663--1676 (2023)

\bibitem{aljundi2019online}
Aljundi, R., Belilovsky, E., Tuytelaars, T., Charlin, L., Caccia, M., Lin, M., Page-Caccia, L.: Online continual learning with maximal interfered retrieval. Advances in neural information processing systems  \textbf{32} (2019)

\bibitem{ashraf2021dogfight}
Ashraf, M.W., Sultani, W., Shah, M.: Dogfight: Detecting drones from drones videos. In: Proceedings of the IEEE/CVF Conference on Computer Vision and Pattern Recognition. pp. 7067--7076 (2021)

\bibitem{cermelli2022modeling}
Cermelli, F., Geraci, A., Fontanel, D., Caputo, B.: Modeling missing annotations for incremental learning in object detection. In: Proceedings of the IEEE/CVF conference on computer vision and pattern recognition. pp. 3700--3710 (2022)

\bibitem{drones6080202}
Chang, Y., Cheng, Y., Murray, J., Huang, S., Shi, G.: The hdin dataset: A real-world indoor uav dataset with multi-task labels for visual-based navigation. Drones  \textbf{6}(8) (2022). \doi{10.3390/drones6080202}

\bibitem{de2024replay}
De~Monte, R., Pezze, D.D., Ceccon, M., Pasti, F., Paissan, F., Farella, E., Susto, G.A., Bellotto, N.: Replay consolidation with label propagation for continual object detection. arXiv preprint arXiv:2409.05650  (2024)

\bibitem{du2018unmannedaerialvehiclebenchmark}
Du, D., Qi, Y., Yu, H., Yang, Y., Duan, K., Li, G., Zhang, W., Huang, Q., Tian, Q.: The unmanned aerial vehicle benchmark: Object detection and tracking. In: Proceedings of the European conference on computer vision (ECCV). pp. 370--386 (2018)

\bibitem{feng2022overcomingcatastrophicforgettingincremental}
Feng, T., Wang, M., Yuan, H.: Overcoming catastrophic forgetting in incremental object detection via elastic response distillation. In: Proceedings of the IEEE/CVF Conference on Computer Vision and Pattern Recognition (CVPR). pp. 9427--9436 (June 2022)

\bibitem{jegham2024yolo}
Jegham, N., Koh, C.Y., Abdelatti, M., Hendawi, A.: Yolo evolution: A comprehensive benchmark and architectural review of yolov12, yolo11, and their previous versions. arXiv preprint arXiv:2411.00201  (2024)

\bibitem{jiang2021anti}
Jiang, N., Wang, K., Peng, X., Yu, X., Wang, Q., Xing, J., Li, G., Zhao, J., Guo, G., Han, Z.: Anti-uav: A large multi-modal benchmark for uav tracking. arXiv preprint arXiv:2101.08466  (2021)

\bibitem{jordan2018state}
Jordan, S., Moore, J., Hovet, S., Box, J., Perry, J., Kirsche, K., Lewis, D., Tse, Z.T.H.: State-of-the-art technologies for uav inspections. IET Radar, Sonar \& Navigation  \textbf{12}(2),  151--164 (2018)

\bibitem{khanam2024yolov11}
Khanam, R., Hussain, M.: Yolov11: An overview of the key architectural enhancements. arXiv preprint arXiv:2410.17725  (2024)

\bibitem{kirkpatrick2017overcoming}
Kirkpatrick, J., Pascanu, R., Rabinowitz, N., Veness, J., Desjardins, G., Rusu, A.A., Milan, K., Quan, J., Ramalho, T., Grabska-Barwinska, A., et~al.: Overcoming catastrophic forgetting in neural networks. Proceedings of the National Academy of Sciences  \textbf{114}(13),  3521--3526 (2017)

\bibitem{kisel2024flawsimagenetcomputervisions}
Kisel, N., Volkov, I., Hanzelkova, K., Janouskova, K., Matas, J.: Flaws of imagenet, computer vision's favourite dataset. arXiv preprint arXiv:2412.00076  (2024)

\bibitem{lin2015microsoftcococommonobjects}
Lin, T.Y., Maire, M., Belongie, S., Hays, J., Perona, P., Ramanan, D., Doll{\'a}r, P., Zitnick, C.L.: Microsoft coco: Common objects in context. In: European conference on computer vision. pp. 740--755. Springer (2014)

\bibitem{liu2024uav}
Liu, H., Tsang, Y.P., Lee, C.K., Wu, C.H.: Uav trajectory planning via viewpoint resampling for autonomous remote inspection of industrial facilities. IEEE Transactions on Industrial Informatics  \textbf{20}(5),  7492--7501 (2024)

\bibitem{liu2023augmented}
Liu, Y., Cong, Y., Goswami, D., Liu, X., Van De~Weijer, J.: Augmented box replay: Overcoming foreground shift for incremental object detection. In: Proceedings of the IEEE/CVF international conference on computer vision. pp. 11367--11377 (2023)

\bibitem{menezes2023continual}
Menezes, A.G., de~Moura, G., Alves, C., de~Carvalho, A.C.: Continual object detection: a review of definitions, strategies, and challenges. Neural networks  \textbf{161},  476--493 (2023)

\bibitem{mo2024bridge}
Mo, Q., Gao, Y., Fu, S., Yan, J., Wu, A., Zheng, W.S.: Bridge past and future: Overcoming information asymmetry in incremental object detection. In: European Conference on Computer Vision. pp. 463--480. Springer (2024)

\bibitem{Mocanu_2025_ICCV}
Mocanu, S., Nae, S.I., Barbu, M.E., Leordeanu, M.: Efficient self-supervised neuro-analytic visual servoing for real-time quadrotor control. In: Proceedings of the IEEE/CVF International Conference on Computer Vision (ICCV) Workshops. pp. 1744--1753 (October 2025)

\bibitem{mueller2016benchmark}
Mueller, M., Smith, N., Ghanem, B.: A benchmark and simulator for uav tracking. In: European conference on computer vision. pp. 445--461. Springer (2016)

\bibitem{neuwirth2025rico}
Neuwirth-Trapp, M., Bieshaar, M., Paudel, D.P., Van~Gool, L.: Rico: Two realistic benchmarks and an in-depth analysis for incremental learning in object detection. In: Proceedings of the IEEE/CVF International Conference on Computer Vision (ICCV) Workshops. pp. 5153--5164 (October 2025)

\bibitem{peng2020faster}
Peng, C., Zhao, K., Lovell, B.C.: Faster ilod: Incremental learning for object detectors based on faster rcnn. Pattern recognition letters  \textbf{140},  109--115 (2020)

\bibitem{reis2023realtime}
Reis, D., Kupec, J., Hong, J., Daoudi, A.: Real-time flying object detection with yolov8. arXiv preprint arXiv:2305.09972  (2023)

\bibitem{ren2024groundedsamassemblingopenworld}
Ren, T., Liu, S., Zeng, A., Lin, J., Li, K., Cao, H., Chen, J., Huang, X., Chen, Y., Yan, F., et~al.: Grounded sam: Assembling open-world models for diverse visual tasks. arXiv preprint arXiv:2401.14159  (2024)

\bibitem{rolnick2019experience}
Rolnick, D., Ahuja, A., Schwarz, J., Lillicrap, T., Wayne, G.: Experience replay for continual learning. Advances in neural information processing systems  \textbf{32} (2019)

\bibitem{selvaraju2020grad}
Selvaraju, R.R., Cogswell, M., Das, A., Vedantam, R., Parikh, D., Batra, D.: Grad-cam: visual explanations from deep networks via gradient-based localization. International journal of computer vision  \textbf{128},  336--359 (2020)

\bibitem{shalaby2025miluvmultiuavindoorlocalization}
Shalaby, M.A., Dahdah, N., Ahmed, S.S., Cossette, C.C., Le~Ny, J., Forbes, J.R.: Miluv: A multi-uav indoor localization dataset with uwb and vision. The International Journal of Robotics Research p. 02783649251405898 (2025)

\bibitem{tian2019fcosfullyconvolutionalonestage}
Tian, Z., Shen, C., Chen, H., He, T.: Fcos: Fully convolutional one-stage object detection. In: Proceedings of the IEEE/CVF International Conference on Computer Vision (ICCV) (October 2019)

\bibitem{wang2021wanderlust}
Wang, J., Wang, X., Shang-Guan, Y., Gupta, A.: Wanderlust: Online continual object detection in the real world. In: Proceedings of the IEEE/CVF international conference on computer vision. pp. 10829--10838 (2021)

\bibitem{wang2024rtdetrv3realtimeendtoendobject}
Wang, S., Xia, C., Lv, F., Shi, Y.: Rt-detrv3: Real-time end-to-end object detection with hierarchical dense positive supervision. In: 2025 IEEE/CVF Winter Conference on Applications of Computer Vision (WACV). pp. 1628--1636. IEEE (2025)

\bibitem{wen2018unmanned}
Wen, F., Wolling, J., McSweeney, K., Gu, H.: Unmanned aerial vehicles for survey of marine and offshore structures: a classification organization's viewpoint and experience. In: Offshore Technology Conference. p. D041S046R003. OTC (2018)

\bibitem{zhu2018visionmeetsdroneschallenge}
Zhu, P., Wen, L., Bian, X., Ling, H., Hu, Q.: Vision meets drones: A challenge. arXiv preprint arXiv:1804.07437  (2018)

\end{thebibliography}

\end{document}